# Classifying Mental-Disorders through Clinicians' Subjective Approach based on Three-way Decisions

Huidong Wang, Md Sakib Ullah Sourav, Mengdi Yang, Jiaping Zhang

*School of Management Science and Engineering, Shandong University of Finance and Engineering, Jinan, China*

A B S T R A C T

In psychiatric diagnosis, a contemporary data-driven, manual-based method for mental disorders classification is the most popular technique. However, it has several inevitable flaws, namely, misdiagnosis of a complex phenomenon, comorbidities etc. Using the three-way decisions (3WD) as a framework, we propose a unified model that stands for clinicians' subjective approach (CSA) consisting of three parts: quantitative analysis, quantitative analysis, and evaluation-based analysis. A ranking list and a set of numerical weights based on illness magnitude levels according to the clinician's greatest degree of assumptions are the findings of the qualitative and quantitative investigation. We further create a comparative classification of illnesses into three groups with varying important levels; a three-way evaluation-based model is utilized in this study for the aim of understanding and portraying these results in a more clear way. This proposed method might be integrated with the manual-based process as a complementary tool to improve precision while diagnosing mental disorders.

## 1. Introduction

In this modern era, where technology is at its peak, with endless amusement and entertainment scopes, still, a substantial amount of people, mostly young adults, are suffering from depression and other mental disorders [1]. Prevalence can be seen as having a lack of motivation to live, losing interest in everything among common people. Hence, they are frequently thriving towards psychiatric diagnosis than in the past days. Therefore, improper diagnosis of mental health disorders may lead to even more vulnerable consequences in a greater sense from an individual to a social perspective [38]. The traditional form of psychiatric diagnosis is much pretentious nowadays as few recent studies demonstrate several shortcomings within the widely established systems used for classifying mental disorders, namely, bipolar disorder, anxiety disorders, phobias, substance use disorder, mood disorders, and many others [2,3]. More often these recognized tools, such as DSM-5 [7] and ICD-11 [8], fails to distinguish between the proper and correct disorder diagnosis of a complex phenomenon in individual cases. Patients with the same disorder exhibit diverse symptom profiles during diagnosis [35] and comorbidities or co-occurring conditions creating numerous clinical and research challenges as well [36]. In such a situation, the pragmatic and expertise-oriented judgment from the clinicians (psychiatrists and psychologists) should be reinforced to avoid an improper diagnosis of a mental disorder and restrict its consequences. While three-way classification has emerged as a prominent problem-solving and decision-making paradigm, we intend to integrate its theory into the classification process of mental disorders in order to help the clinicians' diagnosis process in a more accurate and confident manner.

"Psychiatric nosology" or "psychiatric taxonomy" is terms used to describe how mental diseases are classified. There are presently two commonly used instruments or methods for defining mental disorders: the World Health Organization's (WHO) International Classification of Diseases (ICD-11) and the American Psychiatric Association's (APA) Diagnostic and Statistical Manual of Mental Disorders (DSM-5). In contrast to the American Psychiatric Association's (APA) Diagnostic and Statistical Manual of Mental Disorders (DSM), Research Domain Criteria (RDoC) was launched in 2009 with the goal of addressing the heterogeneity in current nosology by providing a biologically-based, rather than symptom-based, a framework for understanding mental disorders [33]. The Chinese Society of Psychiatry (CSP) [9] produced the Chinese Classification of Mental Diseases (CCMD), a clinical reference for diagnosing mental disorders in China. It is now working on the CCMD-3, a third edition was written in both Chinese and English. It is designed to be structurally and categorically identical to the International Classification of Diseases (ICD) and the Diagnostic and Statistical Manual (DSM).

One of the most fundamental flaws in the DSM-5 and other manuals is that they lack culture-specific meaning and do not include the cultural context of a certain nation (for example, Bangladesh). Common people's habits, tastes, life expectations, social behavior is much more distinct and unique in different parts of the world and these changes rapidly. After the emergence of COVID-19 amidst the imposed restrictions of various sorts, the mental health circumstances is in big threat; the symptoms are relapsing in normal population, university students, clinical workers, patients with pre-psychiatric disorders, and others in such a way that makes the situation more complex [39, 40, 41]. In addition, these taxonomies or guides are mostly based on various statistical analyses and information theory, with some cultural representations thrown in for good measure yet backfire to provide us a timely, holistic and unified view on a deeper scale. On the other hand, the broadening of diagnostic criteria in DSM-5, according to critics, may increase the number of 'mentally ill' people and/or pathologies 'normal behavior', thus exposing millions of additional patients to pharmaceuticals that may do more damage than good [4]. What is more, the different manual-guided psychiatric diagnoses follow approaches like- categorical, dimensional, and others, those also have their controversy in terms of their validity in many cases [5, 6].

Prior to the introduction of manual-based diagnostic systems (about 1980), the clinician's subjective experience was highly regarded. Although the effectiveness of the method may have increased since the DSM/ICD was introduced, the limitations of this technique are now evident [2, 3, 5, 6, 42, and 43]. A study [44] on clinician's subjective experience supports the resurrection of growing potential on clinician's subjectivity and its promising role in diagnostic process. Other recent studies [39, 40] show evidence that the clinician's subjective experience might play a useful role in the diagnostic process as well.

The term "diagnosis" refers to both a phrase and a procedure that is closely linked to concerns of classification [45]. In conventional psychiatric diagnosis process, a doctor or clinician classify among listed mental disorders by referring to the outlined and data-driven manuals (DSM-5/ ICD-11) that include descriptions, symptoms, and so forth; and by following other diagnostic criteria. This is an objective approach that implies internal information-based analysis and it has been put much of the importance comparatively. Simultaneously, similar importance should be imposed on practitioners' external analysis, namely, culture-specific knowledge along with domain knowledge, through attained expertise and experience with a subjective approach during diagnosis process that has been focused on in the current study, this is shown in Fig.1.

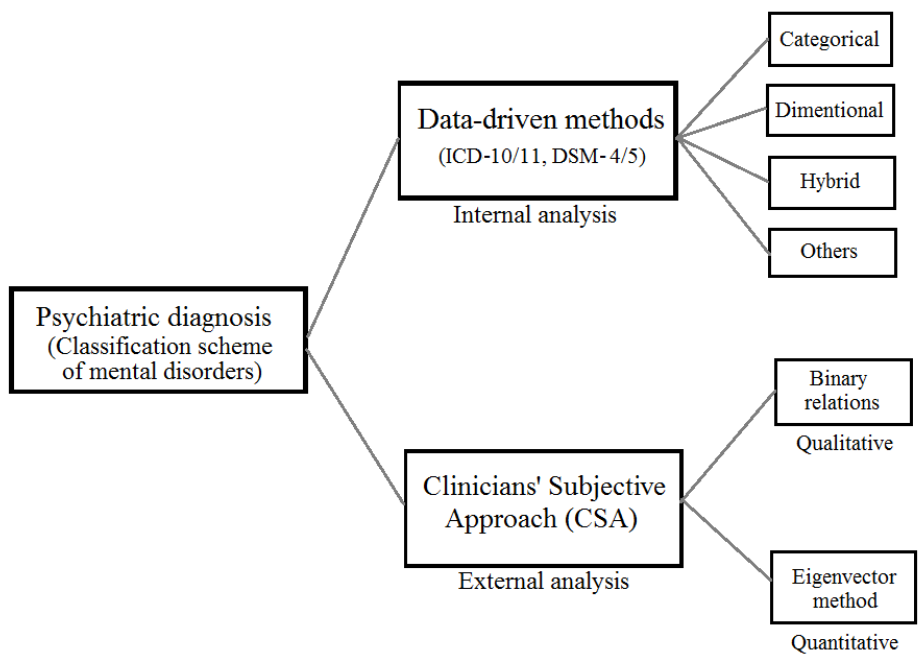

**Fig.1.** A unified framework of mental disorder classification consisting clinicians' subjective approach (CSA).

The primary purpose of this paper is to provide a general framework and adopt concrete methods to analyze clinicians' subjective approach (CSA) in mental disorder classification or psychiatric nosology. The content of this paper is generally arranged into three parts, qualitative analysis of CSA using binary relations, quantitative analysis of CSA using the eigenvector method, and evaluation-based analysis.

## 2. Three-Way Decision in Clinicians' Subjective Approach (CSA)

Yao [10] created the three-way decisions (3WD) theory, which tries to provide a cohesive framework for thinking, problem-solving, and information processing in three dimensions. It gives us a useful foundation for simulating real-world challenges. 3WD has been applied to a variety of fields, including three-way conflict analysis [11,12], three-way clustering [13,14,15,16], three-way recommender systems [17,18,19], three-way concept analysis [20,21], three-way granular computing [10, 22], three-way face recognition [24], and so on.

We use 3WD theory as our fundamental framework to examine clinicians' subjective approach (CSA) to classify mental disorders in psychiatric diagnosis. CSA is studied using three models: the Trisecting-Acting-Outcome (TAO) model, the three-level computing model, and the evaluation-based approach. As shown in Fig. 1, we undertake a qualitative and quantitative investigation of CSA.

We use CSA to rate a group of mental disorders in the qualitative study. The ranking is based on the order relationships between illness pairings. To model the structure and assess CSA, we apply TAO mode in three-way choices. From the standpoint of a clinician, we compare and assess the relative preference of all disorders in pairs first. Following that, we divide all of these couples into three categories: preferred, neutral, and less favored. Finally, we rank illnesses/disorders in order by using a binary relationship. The eigenvector approach is used in quantitative analysis to calculate disease weights. A disadvantage of the eigenvector technique is that when the number of items exceeds 9, a large mistake in the computation might occur [25]. By constructing a three-level structure, the three-level computing paradigm is employed to solve this problem. The eigenvector approach is then used to calculate weights from top to bottom numerous times, allowing us to get a large number of disease weights without sacrificing too much precision.

The findings of the qualitative and quantitative study are a ranking list and a set of numerical weights based on the magnitude levels of illnesses based on the clinician's most extreme assumptions. We also built a comparative classification of illnesses into three groups with varying important levels, three-way evaluation-based model is utilized in this study for the aim of comprehending and portraying these results in a more straightforward way.

## 3. Three-Way Qualitative Clinicians' Subjective Approach (CSA) Analysis

Order relations, which is an intuitive sense of ordering things against one another, are a significant implication of binary relations. For example, given that (x, y) is an ordered pair of two elements, we may derive order relations between x and y, such as x being greater than y, x being poorer than y, or x being a component of y in various instances. Order relations are a frequent representation of user preference in decision theory, we write an order relation as $\geq$ or $>$. If $x \geq y$, we say x is at least as good as y, if $x > y$, we say x is strictly better than y. We solely focus on the strict order relation of "$>$" in this study to develop a clinician's preference-based approach (CPA) later on in a more clear way based on the property of trichotomy.

*3.1. Clinicians' Subjective Approach (CSA) and the Property of Trichotomy*

The idea of user preference has been intensively investigated in several user-oriented research domains, such as information retrieval [26, 27], economics [28], and social sciences [29]. In qualitative CSA analysis, the concept of user preference theory may be employed, and the feature of trichotomy is crucial. This trait makes order relations useful for modeling a CSA towards a set of illnesses.

Humans are skilled at establishing relative comparisons between numbers, goods, methods, and other things in our daily lives. Given two arbitrary real numbers n and m, we may easily argue that one of nm, n = m, or n > m must hold in number theory; this is known as the trichotomy property of real numbers. Similarly, a person can identify the ordering relation between x and y as one of the following: x is preferred over y, x is indifferent to y, or x is less favored than y by comparing a pair of things x and y under a specified criterion. Obviously, a person's preferred preference for a pair of things is three. This concept can easily be expanded to include order relations.

If we use an order relation > to represent the meaning "preferred", the indifference relation ~ is defined as an absence of >, which is defined as:

$$x \sim y \Leftrightarrow \neg (x > y) \wedge \neg (y > x) \tag{1}$$

Give an ordered pair (x, y), if an order relation > expresses the first element is preferred than the second element. Its converse relation which is written as $\succ$, is called a less preferred relation, which is defined as:

$$x \succ y \Leftrightarrow (y > x) \tag{2}$$

We usually write $\succeq$ as $\prec$ if it does not cause any ambiguity.

**Definition 1.** *An order relation $\succ$ on a disorder set D is called trichotomous if $\forall(x, y), x, y \in D$, exactly one of $x > y$, $x \sim y$, or $x \prec y$ holds.*

The purpose of user preference-related research, from the perspective of a decision-maker, is to identify optimum options by examining the order relations among members of a nonempty set, which is characterized as preference relation. The method of user preference theory may be described as first establishing reasonable axioms based on the decision maker's preferences, and then assessing a user's preferring behavior based on those preferences [28]. The mathematical properties of trichotomy and transitivity are used to construct a preference relation.

**Definition 2.** A preference relation, denoted as >, is a special type of binary relation on the set of elements D that satisfies the following two rationality properties. $\forall x, y, z \in D$,

$$\text{Trichotomous: } (x > y) \vee (x \sim y) \vee (x \prec y),$$

$$\text{Transitive: } x > y \wedge y > z \Rightarrow x > z \qquad (3)$$

If we use an order relation > as a preference relation, user preference is represented as:

$$\begin{aligned} x > y &\Longleftrightarrow x \text{ is preferred than } y \\ x \sim y &\Longleftrightarrow x \text{ is indifferent with } y \\ x \prec y &\Longleftrightarrow x \text{ is less preferred than } y \end{aligned} \qquad (4)$$

For a disorder set Ds, we divide all disorder attribute pairs into three classes. Based on this trisection, disorder ranking can be induced. This process is shown in Fig. 2:

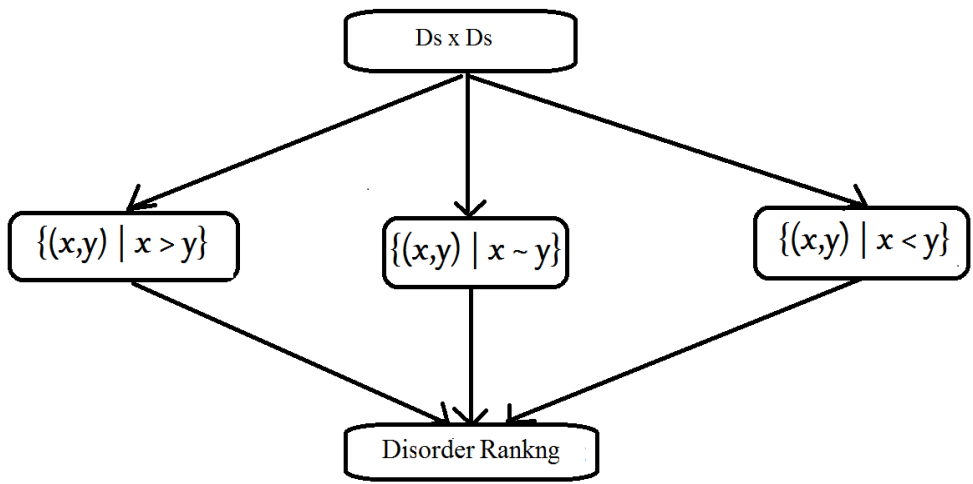

**Fig.2.** The property of trichotomy.

Linear orders, weak orders, and semiorders are the three types of order relations that all have the properties of trichotomy and transitivity. These three order relations are employed in this article to describe the clinician's choice for CSA analysis.

*3.2 Modeling of CSA as Linear Order*

Given a disorder set Ds, a linear order > enables us to arrange diseases in the form Ds = $\{d_1, d_2,..., d_n\}$, such that $d_i > d_j$ if and only if I < j, for this reason, a linear order is also called a chain.

**Definition 3.** Given a set Ds, a binary relation > is a linear order on Ds, if it satisfies for any x, y, z ∈ Ds:

$$\text{Asymmetric: } x > y \Rightarrow \neg (y > x),$$
$$\text{Transitive: } x > y \wedge y > z \Rightarrow x > z,$$
$$\text{Weakly Complete: } x \neq y \Rightarrow (x > y) \vee (y > x) \quad (5)$$

The asymmetric feature precludes the circumstance in which $d_i$ is better than $d_j$ and $d_j$ is better than $d_i$ at the same time. Reasonable inference may be applied thanks to the transitive property. Weak completeness assures that all illnesses are comparable to one another.

**Example 1.** Given a set of disorder Ds = {d1, d2, d3, d4, d5}, a clinicians' preference in accordance of the assumptions on a patient having a potential disorder on Ds is defined by a linear order >. Suppose the ordering between disorders is specified by a clinician as:

$$d1 \succ d5, d1 \succ d4, d1 \succ d2, d3 \succ d1, d3 \succ d2,$$
$$d3 \succ d4, d3 \succ d5, d5 \succ d4, d5 \succ d2, d4 \succ d2.$$

Then, disorders are ranked as:

$$d3 \succ d1 \succ d5 \succ d4 \succ d2.$$

*3.3 Modeling of CSA as Weak Order to Illustrate Comorbidity in Mental Disorder*

Weak orders are commonly utilized in several disciplines to indicate user preference relations [26, 27, 28, and 29]. A weak order enables ties in the ranking results, as opposed to a linear order that places items in a chain, which is quite powerful in representing real-world issues. To put it another way, some properties in a collection may be regarded as indifferent.

In mental disorder classifications, comorbidity of psychiatric illnesses is a widespread issue with major consequences for health-care delivery [36]. Depression, anxiety, and drug dependency disorders are the most common comorbid mental illnesses [37]. Here, we used ~ to denote comorbidity of mental disorders and weak ordered relation > to denote ranking of clinician's preference among disorders.

**Definition 4.** A weak order > is a binary relation on set Ds, if it satisfies for any x, y ∈ Ds:

$$\text{Asymmetric: } x > y \Rightarrow \neg (y > x),$$

$$\text{Negative transitive: } \neg (x > y) \wedge \neg (y > z) \Rightarrow \neg (x > z) \quad (6)$$

**Example 2.** Given a set of disorder Ds = {d1, d2, d3, d4, d5}, a clinician's preference on Ds is defined by a weak order >. Suppose the ordering between disorders is specified as:

$$d1 > d3, d1 > d4, d1 > d5, d2 > d3, d2 > d4, d2 > d5, d3 > d4, d3 > d5.$$

Because the clinician neither preferences d1 to d2, nor prefer d2 to d1, so d1 must be in comorbid condition or indifferent with d2, written d1 ~ d2. That means the clinician suspects the particular patient has disease d1 and d2 at the same time. Similarly, d4 ~ d5. By considering the above ordering, we can rank disorder attributes like:

$$d1 \sim d2 > d3 > d4 \sim d5.$$

*3.4 Modeling of CSA as Semiorder*

In fact, a transitive indifference relationship isn't always the case. A reader may assume that books C and D are equally good, as are books D and E, after reading three novels, yet he can know that he prefers C to E based on his intuition after reading three books. To put it another way, the individual's preferring attitude cannot discriminate neither between C and D, nor between D and E, but he can distinguish between C and E. To model this type of situation, Luce [31] proposed semiorders.

**Definition 5**. A semiorder > on a set Ds is a binary relation which satisfies for any x, x', x'', y, y' ∈ Ds:

$$\begin{aligned}
&\text{Asymmetric: } x > y \Rightarrow \neg (y > x), \\
&\text{Ferrers: } (x > x') \wedge (y > y') \Rightarrow (x > y') \vee (y > x'), \\
&\text{Semi transitive: } (x > x') \wedge (x' > x'') \Rightarrow (x > y) \vee (y > x'')
\end{aligned} \quad (7)$$

**Example 3.** Given a set of disorder Ds = {d1, d2, d3, d4, d5}, a clinician's preference on Ds is defined by a semiorder >. Suppose the ordering between disorders is specified as:

$$d1 > d2, d1 > d3, d1 > d4, d1 > d5, d2 > d4, d2 > d5, d3 > d5, d4 > d5.$$

The clinician neither prefers d2 to d3, nor prefer d3 to d2, so d2 ~ d3, similarly we can get d3 ~ d4, however, the indifference is intransitive, because d2 > d4. So, we cannot rank all disorders in one order but several, like below:

$$d1 > d2 > d4 > d5,$$
$$d1 > d2 \sim d3 > d5,$$
$$d1 > d3 \sim d4 > d5.$$

**4. Three-Way Quantitative Clinicians' Subjective Approach (CSA) Analysis**

Mathematically, quantitative CSA analysis can be considered as a process of mapping each disorder to a numerical value,

$$w: Ds \to R \tag{8}$$

where Ds is a set of disorders, R is a real number set, and w is a mapping function that calculates or assigns a numerical value to each disorder. For a disorder $d \in Ds$, $w(d)$ represents its weight from the perspective of a clinician.

*4.1 Formulating a Three-Level Structure*

This study offered two methods for calculating or assigning numerical weights to each ailment. The first is calculating weights using the eigenvector approach, which is covered in Sect. 4.2. The second method is to assign weights. To be more explicit, we first use the eigenvector method to construct an important scale with numerical weights, and then we compare each disease to this scale to determine its weight; this methodology is detailed in Sect. 4.3. Obviously, the eigenvector technique is vital in both approaches; however, it has a limitation in that it is not suitable when the number of objects is greater than 9 since large mistakes in the computation would be introduced [25]. We use the 3WD theory to solve this problem. More specifically, the issue is divided into three levels, after which the eigenvector approach is used to calculate weights from top to bottom. The three-level structure allows us to limit the number of items in the computation of the weights to no more than 9, allowing us to compute weights using the eigenvector approach without sacrificing too much precision.

*4.2 Three-Way Quantitative Disease Weighting Based on Eigenvector Method*

Figure 3 depicts the framework of the quantitative illness weighting model. Assume we have a disorder set Ds, where $d_{ij}$ denotes a disorder at the lowest level. We create a three-level framework by categorizing illnesses into various groups based on semantic significance.

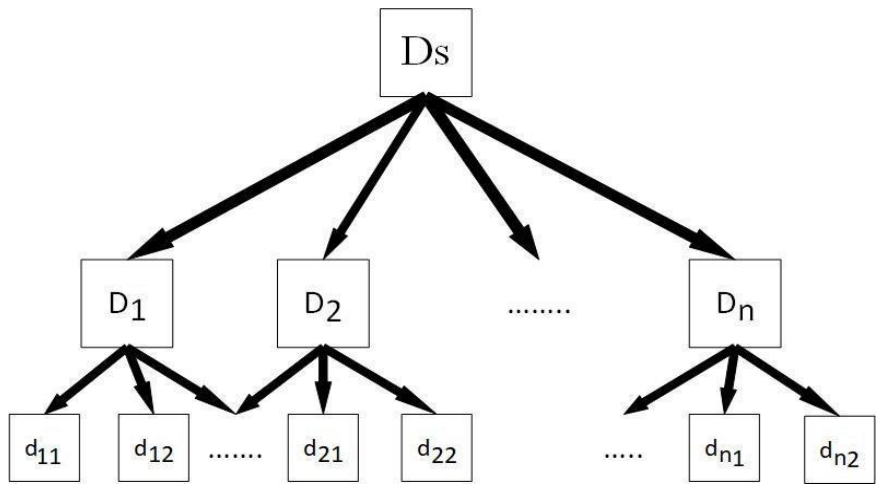

**Fig. 3.** The structure of the three-level disease weighting method.

The second step is to use the eigenvector approach to calculate from top to bottom after we have this three-level structure. We create cluster weights based on clinician choice, and then we determine the weights of illnesses inside each cluster based on cluster weight.

The following is a description of how to calculate weights using the eigenvector approach. Assume that a disorder collection Ds has been divided into n clusters, n 9, with no more than 9 illnesses in each cluster. We establish a comparison matrix M as specified in Definition 6 to produce a weight vector w = ($w_1$, $w_2$, ⋯ ,$w_n$) for clusters, where element $m_{ij}$ reflects the relative significance of a cluster Di compared to a cluster Dj.

**Definition 6**. A comparison matrix M is a square matrix of order n, whose elements are $m_{ij}$, M is a positive reciprocal matrix if M is:

$$\text{Positive: } \forall i, j < n, m_{ij} > 0,$$
$$\text{Reciprocal: } \forall i, j < n, m_{ij} = 1/m_{ji}. \tag{9}$$

Where, $\forall i, j$ (i, j=1, 2… n).

M is a comparison matrix that looks like below, and in a perfect situation, $m_{ij}$ should exactly be the weights ratio of a cluster Di compared with $D_j$.

$$M = \begin{pmatrix} m_{11} & m_{12} & m_{1n} \\ m_{21} & m_{22} & m_{2n} \\ . & . & . \\ . & . & . \\ . & . & . \\ m_{n1} & m_{n2} & m_{nn} \end{pmatrix} = \begin{pmatrix} w_1/w_1 & w_1/w_2 & ...w_1/w_n \\ w_2/w_1 & w_2/w_2 & ...w_2/w_n \\ . & . & . \\ . & . & . \\ . & . & . \\ w_n/w_1 & w_n/w_2 & ...w_n/w_n \end{pmatrix} \tag{10}$$

In practice, the values of components in a comparison matrix are determined by the user's preference and flexibility. We use the 9-point rating scale established by Saaty [25] to precisely determine the weight ratio w1/w2 between two clusters (see Table 1).

**Table 1.** The Saaty's 9-points rating scale [25]

| Intensity of Importance | Definition | Explanation |
|---|---|---|
| 1 | Equal importance | Two activities contribute equally to the objective |
| 3 | Weak importance of one over another | Experience and judgment slightly favor one activity over another |
| 5 | Essential or strong importance | Experience and judgment strongly favor one activity over another |
| 7 | Demonstrated importance | An activity is strongly favored and its dominance demonstrated in practice |
| 9 | Absolute importance | The evidence favoring one activity over another is of the highest possible order of affirmation |
| 2,4,6,8 | Intermediate values between the two adjacent judgments | When compromise is needed |

Table 1 shows that the number 1 denotes that two clusters are equally essential. An arbitrary cluster should be equally significant to itself, hence the value mii of the major diagonal in a comparison matrix must be 1. Furthermore, for two clusters a and b, the weight ratio wa/wb should be larger than 1 if an is favored over b, else it should be equal to or less than 1.

We may get the matrix equation as follows under ideal conditions:

$$Mw = \begin{pmatrix} w_1/w_1 & w_1/w_2 & ...w_1/w_n \\ w_2/w_1 & w_2/w_2 & ...w_2/w_n \\ . & . & . \\ . & . & . \\ . & . & . \\ w_n/w_1 & w_n/w_2 & ...w_n/w_n \end{pmatrix} \times \begin{pmatrix} w_1 \\ w_2 \\ . \\ . \\ . \\ w_n \end{pmatrix} = nw \qquad (11)$$

Where, M is multiplied on the left by the weights vector w, and the result is nw. The issue we're working on has been rewritten as Mw = nw = 0, or (M nI)w = 0. Ideally, M is consistent if and only if its principle eigenvalue $\lambda_{max}$ = n [25]. Inconsistency is unavoidable since items in a comparison matrix are personal judgments evaluated by a clinician. Precisely because of this, changes in the matrix lead to changes in the eigenvalues. We now have a new challenge to solve:

$$M'w' = \lambda_{max} w', \qquad (12)$$

Where $M' = (m'_{ij})$ is the perturbed matrix of M = $(m_{ij})$, $w'$ is the principal eigenvector and $\lambda_{max}$ is the principal eigenvalue of $M'$. What we want to learn is how good the principal eigenvector $w'$ represents w. Consistency ratio C.R. is used to determine whether an inconsistency is acceptable:

$$C.R. = \frac{\lambda max - n}{(n-1) \times R.I.} \qquad (13)$$

Table 2 shows the average random consistency index (R.I.). These indices were created using a 9-point rating scale and a sample of randomly generated reciprocal matrices [25].

**Table 2**. Average random consistency index (R.I.)

| n   | 1 | 2 | 3    | 4    | 5    | 6    | 7    | 8    | 9    | 10   |
|-----|---|---|------|------|------|------|------|------|------|------|
| R.I | 0 | 0 | 0.52 | 0.89 | 1.11 | 1.25 | 1.35 | 1.40 | 1.45 | 1.49 |

Eigenvector w can be used as cluster weights if C.R. is less than 10%; otherwise, the comparison matrix must be changed until C.R. is less than 10%.

**Example 4.** Assume we have a disorder set Ds that has been divided into six clusters, Ds = {D1, D2, D3, D4, D5, D6}. Based on a clinician's assessment, a comparison matrix has been created, and the weights calculation procedure is illustrated in Table 3 [34]:

Table 3. Weights calculation of six clusters

|    | D1  | D2 | D3  | D4 | D5  | D6  | Weight |
|----|-----|----|-----|----|-----|-----|--------|
| D1 | 1   | 3  | 1/2 | 4  | 2   | 1/3 | 0.140  |
| D2 | 1/3 | 1  | 1/7 | 1  | ½   | 1/9 | 0.041  |
| D3 | 2   | 7  | 1   | 9  | 5   | ½   | 0.290  |
| D4 | ¼   | 1  | 1/9 | 1  | ½   | 1/9 | 0.038  |
| D5 | ½   | 2  | 1/5 | 2  | 1   | 1/6 | 0.071  |
| D6 | 3   | 9  | 2   | 9  | 6   | 1   | 0.420  |

$\lambda max$= 6.048

C.R = 0.762% < 10%

Because C.R.≤ 10%, which satisfies consistency checking, the eigenvector of comparison matrix can be used as the weights of {D1, D2, D3, D4, D5, D6}, that is:

w = (0.140, 0.041, 0.290, 0.038, 0.071, 0.420)

We utilize the weights of clusters as a starting point and use the same method to compute the weights of illnesses in each cluster. The weights of illnesses are then normalized by dividing them by the weights of respective clusters. Finally, we may calculate illness weights.

*4.3 A Quantitative Disease Weighting Method Using an Importance Scale*

It's a simple approach for a doctor to assign numerical numbers to illnesses as weights depending on his or her own viewpoint. When the number of suspected illnesses is enormous, however, fluctuation in judgment is unavoidable, resulting in low accuracy in the conclusion. In light of this, an importance scale is employed to solve the problem [25].

The disease weighting approach employing a significance scale may be broken down into the three parts below. First, intensities of the preference degree of illness orders are grouped into several degrees from a clinician's perspective, such as significantly matched, matched, moderately matched, weakly matched, and not matched. We can then calculate weights for each intensity degree using the eigenvector approach described in Sect. 4.2. A three-level structure is required when the number of intensity degrees exceeds 9. As a result, we create a importance scale to aid our judgment. Finally, the weights of illnesses are calculated using this scale.

**Example 5.** Suppose a clinician sets five intensities of the preferential degree of suspected disorders, which are A: significantly matched, B: matched, C: moderately matched, D: weakly matched, E: not matched. A clinician builds a comparison matrix of these intensities, and the weights calculation of intensities is described as (Table 4) [34]:

Table 4. A pairwise comparison matrix of intensity levels

|   | A   | B   | C   | D   | E | Weight |
|---|-----|-----|-----|-----|---|--------|
| A | 1   | 2   | 3   | 5   | 9 | 0.450  |
| B | ½   | 1   | 2   | 4   | 6 | 0.277  |
| C | 1/3 | ½   | 1   | 2   | 3 | 0.147  |
| D | 1/5 | ¼   | ½   | 1   | 2 | 0.081  |
| E | 1/9 | 1/6 | 1/3 | ½   | 1 | 0.046  |

$\lambda max$= 5.024

C.R = 0.533%

Because the consistency check is complete, the weights of these intensities are used to construct an important scale. Then, using this scale, we compare each property one by one, assigning various weights to each disorder attribute from the clinician's perspective.

**5. Three-Way Evaluation based CSA Analysis**

The 3WD [30] is based on dividing the universe into three zones and employing various tactics in each. The result of a qualitative or quantitative CPA analysis is a ranking list or a set of numerical weights that are significant but difficult for a physician to use in making a choice. These findings will be processed and classified into three pair-wise disjoint classes with varying levels of relevance, namely high importance, medium importance, and low importance, in this part. We'll refer to these three classes as H, M, and L for the rest of this study. We chose three classes because human cognition and problem-solving rely on a three-way divide, which allows us to convert complexity into simplicity in a variety of scenarios [23].

*5.1 Trisecting a Disorder Set based on Thresholds*

Research Domain Criteria (RDoC), considering the possibility of increasing need of constructing various thresholds for different purposes, is trying to gather information consistently to set thresholds in diagnostic systems of mental disorder where this is relevant; especially in particular research purpose or applications in clinical settings or in health policymaking [33]. Our current study acknowledges this concern and suggests insightful paths on formulating thresholds in mental disorder classification while in the diagnosis process.

Using two percentiles is one method for trisecting a disorder set. The first step is to convert a linear order > from a qualitative or quantitative analytical result. This phase can be bypassed if the outcome of the qualitative method is based on linear order. To identify the three areas, the second step is to use a pair of thresholds, based on the percentiles.

There are various methods for linearly transforming qualitative and quantitative findings. The first is topological sorting, which states that an element will not appear in a ranking list until all other items that are preferable to it have been listed [32]. We can generate a decreasing ranking list by utilizing topological sorting. Another option is to use an assessment function to convert qualitative and quantitative analytical results into a set of diseases' evaluation status values (ESVs). The ESV of disease d is defined as follows:

$$v(d) = \frac{|\{x \in Ds | d > x\}|}{|Ds|} \quad (14)$$

Illnesses will be sorted in decreasing order depending on their ESVs, with diseases with the same ESV being listed in any order.

Now, we have a list of ESVs, which is in the form of $v_1, v_2..., v_n$ where $v_1$ is the largest value and $v_n$ is the smallest value. Using the ranking lists of the above two methods, we then adopt two ESVs at $\alpha^{th}$ and $\beta^{th}$ percentiles with $0 < \beta < \alpha < 100$ to calculate a pair of thresholds $h$ and $l$ as:

$$h = v_{[\alpha n/100]}$$
$$l = v_{[\beta n/100]} \quad (15)$$

Where the ceiling function $\lceil x \rceil$ gives the smallest integer that is not less than x, and the floor function $\lfloor x \rfloor$ gives the largest integer that not greater than x. The floor and ceiling functions are necessary for the reason that αn/100 and βn/100 may not be integers [30].

Three regions, H, M, and L, may be created using the descending ranking list and two thresholds. Disorders in the H region are of high priority, disorders in the M region are of moderate priority, and disorders in the L zone are of low priority.

*5.2 Trisecting a Disease Set Based on a Statistical Method*

Yao and Gao [30] examined the statistical procedure of building and evaluating three areas. Mean and standard deviation are statistical methods for examining numerical numbers that may be applied to the findings of a quantitative CPA study. Suppose $w(d_1), w(d_2), ..., w(dn)$ are the weights of disorders in Ds, n is the cardinality of Ds, the mean and standard deviation is calculated by:

$$\mu = \frac{1}{n}\sum_{i=1}^{n} w(a_i), \tag{16}$$

$$\sigma = \left(\frac{1}{n}\sum_{i=1}^{n}(w(a_i)-\mu)^2\right)^{\frac{1}{2}}, \tag{17}$$

We use two non-negative numbers $k_1$ and $k_2$ to represent the position of thresholds away from the mean, then a pair of thresholds is determined as [30]:

$$h = \mu + k_1\sigma, \; k_1 \geq 0,$$
$$l = \mu - k_2\sigma, \; k_2 \geq 0. \tag{18}$$

Based on thresholds h and l, three regions of a disorder set can be constructed as:

$$\begin{aligned}
H_{(k1,k2)}(w) &= \{x \in Ds \,|w(x) \geq h\} \\
&= \{x \in Ds \,|w(x) \geq \mu + k_1\sigma\}, \\
M_{(k1,k2)}(w) &= \{x \in Ds \,|l < w(x) < h\} \\
&= \{x \in Ds | \mu - k_2\sigma < w(x) < \mu + k_1\sigma\}, \\
L_{(k1,k2)}(w) &= \{x \in Ds \,|w(x) \leq l\} \\
&= \{x \in Ds \,|w(x) \leq \mu - k_2\sigma\}
\end{aligned} \tag{19}$$

Disorders can be categorized into three regions H, M, and L considering their weights.

**6. Discussion and Future Directions**

We essentially emphasized that the mental symptoms and indicators are passively observed subjects in the paradigm that this research proposes. However, in terms of phenomenological psychopathology, clinicians can also examine patient symptoms and signs using emphatic techniques [46]. Additionally, mental disorders are defined with an operational manner in mainstream diagnostic systems (such as ICD-11 and DSM-5) but are not based on biological indicators. The psychiatric diagnoses therefore correspond to the practical or fuzzy kinds but not the natural kinds. What is more, the operational definitions are relevant to language games in terms of Wittgenstein's philosophy of language [47]. Specifically, the instances with a single diagnosis might be linked by a chain of meanings rather than being supported by a single biological foundation in such disease essentialism. With the addition of psychiatrist-patient interaction (i.e., psychiatrist's emphatic approach) and/or its influences on the classification as future work or extension of this current study, the proposed classification model of mental disorders through clinicians' subjective approach on 3WD can be further modified.

Different paradigms, such as pointing graphs, can be used to develop the disorder ranking procedure. To calculate the weight value of each disorder cluster, analytic hierarchy process can be adopted in lieu of eigenvector method and the results can be compared. The quantification of the weight values for diseases using a three-level structure is required in future by clinicians when the number of intensity degrees exceeds 9. For the time being, this current study is offering the theoretic approach to solve this complex problem related to psychiatric diagnosis. Practical implications in both the qualitative and quantitative perspectives should be explored in further studies to ensure the proposed method is better than other existing methods.

## 7. Conclusions

The most widely used method of mental disorder classification is a data-driven manual-based method, yet it has a number of drawbacks. We offer a three-part unified model for clinicians' subjective approach (CSA) analysis, based on the three-way choice. In the qualitative research, we use binary relations and the TAO model to rank mental diseases based on their criteria-based preferences. The quantitative analysis employs the three-level computing paradigm, and the eigenvector technique is utilized to assign numerical weights to mental diseases. Finally, we categorize the results of the qualitative and quantitative investigations into three groups depending on their relative importance.